# *VUSFA*: <u>V</u>ariational <u>U</u>niversal <u>S</u>uccessor <u>F</u>eatures <u>A</u>pproximator to Improve Transfer DRL for Target Driven Visual Navigation

Shamane Siriwardhana, Rivindu Weerasakera, Denys J.C. Matthies, Suranga Nanayakkara

Augmented Human Lab, Auckland Bioengineering Institute, The University of Auckland, NZ

<firstname>@ahlab.org

## Abstract

In this paper, we show how novel transfer reinforcement learning techniques can be applied to the complex task of target driven navigation using the photorealistic AI2THOR simulator. Specifically, we build on the concept of Universal Successor Features with an A3C agent. We introduce the novel architectural [1] contribution of a Successor Feature Dependant Policy (SFDP) and adopt the concept of Variational Information Bottlenecks to achieve state of the art performance. *VUSFA*, our final architecture, is a straightforward approach that can be implemented using our open source repository. Our approach is generalizable, showed greater stability in training, and outperformed recent approaches in terms of transfer learning ability.

## 1 Introduction

The human's ability of navigating unknown spaces (e.g. a firefighter finding the fire hydrant very quickly) primarily relies on visual perception, as well as on previous experience and heavy training [Ramirez et al., 2009]. In robotics, we would like to mimic this human behaviour. The advancement of visual navigation algorithms essentially contribute to the prevalence and mobility in robotics and therefore, many different approaches are being explored. Previous research has studied map-based, map-building, and map-less approaches [Bonin-Font et al., 2008, Oriolo et al., 1995, Borenstein and Koren, 1991]. In the past, map-based and map-building approaches have been favoured. However, they heavily depend on an accurate mapping of the environment. Also, it requires a carefully executed human-guided training phase which limits its generalizability [Filliat and Meyer, 2003]. With recent advances in Deep Reinforcement Learning (DRL) [Mnih et al., 2015, Silver et al., 2016, 2017], map-less navigation has experienced major advancements [Zhu et al., 2017, Mirowski et al., 2018]. It has been demonstrated that DRL-based methods are now able to solve navigation tasks in a more human-like manner[Fan et al., 2018].

Research has shown that DRL-based navigation, in particular target driven visual navigation, is still a challenging task especially when targets are represented in the form of visual information that is highly dynamic. In previous navigation paradigms, the agent navigates to a target demonstrating specific properties (e.g. a yellow cone, such as in the case of Zhang et al. [2017]), whose location may change over time. In contrast, in target driven visual navigation, the agent should be able to learn to navigate in a persistent state space to a dynamic set of goals. The agent is required to learn to navigate when both the goal and the current state are presented as visual images.

A current challenge for DRL algorithms is learning new tasks or goals that vary from what the agent was initially trained for. This ability is called transfer learning. There are two popular strategies for achieving transfer learning in DRL, either by using the concept of General Value Functions (GVF) [Sutton et al., 2011] or by using Successor Feature Approximation (SFA) [Dayan, 1993]. For the task of target driven visual navigation, Zhu et al. [2017] demonstrated that an A3C agent using the concept of GVF can improve the transfer learning ability. GVF does not however allow us to easily see the underlining process of learning the dynamics of tasks and GVF agents also frequently struggle in complex environments [Sutton et al., 2018]. The second strategy, applying SFA, enables us to capture the dynamics of the environment by attempting to learn future state visitations, although these also encounter limitations when facing multiple tasks. Universal Successor Features Approximators (USFA)[Borsa et al., 2018], which is an extension of SFA, is able to consider multiple tasks and can improve the transfer learning ability of the agent.

---

[1] https://github.com/shamanez/VUSFA-Variational-Universal-Successor-Features-Approximator

In summary, our research contribution is threefold:

- For the first time in the literature, we apply Universal Successor Feature Approximators (USFA) for the complex task of target driven visual navigation. Our new approach provides a stable training mechanism and enhances the transfer reinforcement learning ability in complex environments.

- We introduce the concept of a Successor Feature Dependant Policy (SFDP), a novel architectural contribution in which the policy can directly make use of the information presented by USFA (an abstract map in our case). This important add-on significantly improves the transfer learning ability of the DRL agent.

- Finally, we contribute Variational Universal Successor Feature Approximators (*VUSFA*), by adopting the concept of Variational Information Bottlenecks. We show that this combination works stably with complex tasks such as target driven visual navigation in the photo-realistic AI2THOR environment [Kolve et al., 2017]. Besides stable convergence, our approach shows possible ways in which transfer learning could be improved in the future.

## 2 Background

### 2.1 Transfer in Reinforcement Learning

Transfer in reinforcement learning agents can be described as the ability of an agent to generalize over different tasks while sharing knowledge between them. In this paper we tackle the problem of transfer reinforcement learning with respect to Successor Feature Reinforcement Learning (SF-RL) [Barreto et al., 2017]. Specifically we develop on the concept of Universal Successor Feature RL (USF-RL) [Ma et al., 2018b] which is an extension of SF-RL. In the next section we will first introduce basic concepts where we have used throughout our research.

#### 2.1.1 General Value Functions

We can formalize the goal-directed navigation task as a Markov Decision Process (MDP). The transition probability $p(s_{t+1}|s_t, a_t)$ defines the probability of reaching the next state $s_{t+1}$ when action $a_t \in \mathcal{A}$ is taken in state $s_t \in \mathcal{S}$. For any goal $g \in \mathcal{G}$ (in our case $\mathcal{G} \subseteq \mathcal{S}$), we define a goal dependent reward function $r_g(s_t, a_t, s_{t+1}) \in \mathbb{R}$ and a discount function $\gamma_g(s_t) \in [0, 1]$ (for terminal state, $\gamma_g = 0$). For any policy $\pi(a_t|s_t)$, a GVF [Sutton et al., 2011, Schaul et al., 2015] can be defined as follows:

$$V_g^\pi(s) = \mathbb{E}^\pi \left[ \sum_{t=0}^\infty r_g(s_t, a_t, s_{t+1}) \prod_{k=0}^t \gamma_g(s_k) \middle| s_0 = s \right] \quad (1)$$

The assumption for any goal $g$ is that there exists an optimal value function $V_g^{\pi_g^*}(s)$, which is evaluated according to a goal oriented optimal policy $\pi_g^*$. The general aim of agent's learning is to find the optimal policy $\pi^*$ that maximises the future discounted rewards starting from $s_0$ and following $\pi^*$.

To generalize over the goal space $\mathcal{G}$, the agent needs to learn multiple optimal policies as well as optimal value functions in order to navigate to a goal. Each goal is considered a new task and the agent should be able to quickly adapt to find $V_g^{\pi_g^*}(s)$ and $\pi_g^*$.

#### 2.1.2 Universal Successor Features

Universal Successor Features (USF) [Ma et al., 2018b] in an extension of the idea of Successor Features (SF) described in Kulkarni et al. [2016] and Barreto et al. [2017]. Similar to the concept of SF, USF also follows the idea that the immediate scalar reward $r_g$ can be defined as a linear combination of state representations $\phi$ and a goal dependent reward prediction vector $\omega_g$ as in Equation 2.

In the Equation 2, $\phi(s_t, a_t, s_{t+1})$ represents the dynamics or the physical features the agent sees when transitioning between states $s_t$ and $s_{t+1}$ after taking an action $a_t$. We approximate $\phi(s_t, a_t, s_{t+1})$ as $\phi(s_{t+1})$ following Ma et al. [2018a], Borsa et al. [2018] since it is convenient for the agent to rely on the state representation of the new state $\phi(s_{t+1})$ to recover the scalar reward $r_g$ rather than trying to capture physical features of transition dynamics.

$$r_g(s_t, a_t, s_{t+1}) \approx \phi(s_t, a_t, s_{t+1})^\top \omega_g \\ \approx \phi(s_{t+1})^\top \omega_g \qquad (2)$$

This allows us to describe the value function as a cumulative sum of the discounted $\phi$ as follows:

$$V_g^\pi(s) = \mathbb{E}^\pi \left[ \sum_{t=0}^{\infty} \phi(s_{t+1}) \prod_{k=0}^{t} \gamma_g(s_k) \bigg| s_0 = s \right]^\top \omega_g \\ = \psi_g^\pi(s_t)^\top \omega_g \qquad (3)$$

where $\psi_g^\pi(s_t)$ is defined as the Universal Successor Features (USF) of state $s_t$ [Ma et al., 2018b]. Intuitively, $\psi_g^\pi(s_t)$ can be thought of as the expected future state occupancy. Unlike traditional Successor Feature Approximation, USF is based on both the state and the goal. The value function defined with USFA has similar properties to GVFs. The modified $V_g^\pi(s)$ with $\psi$ incorporates shared dynamics between tasks.

Learning the USFA is accomplished in the same way as the value function update by using the following TD (Temporal Difference) error:

$$\mathcal{L} = \mathbb{E}^\pi[\phi(s_{t+1}) + \gamma_g(s)\psi_g^\pi(s_{t+1})] - \psi_g^\pi(s_t) \qquad (4)$$

As illustrated by [Ma et al., 2018b] the vectors $\psi_g^\pi(s_t)$, $\phi(s_{t+1})$ and $\omega_g$ can be approximated by neural networks parameterized by $\theta_\pi$, $\theta_\phi$ and $\theta_\omega$. In this paper we incorporate the concept of USF with an A3C agent [Mnih et al., 2015] and trained all three sets of parameters jointly.

## 2.2 Applicability of USFA in Large Scale DRL Problems

The USFA model introduced by Ma et al. [2018b] extends the concept of SF [Barreto et al., 2017, Kulkarni et al., 2016] generalized over multiple tasks with the actor-critic algorithm. However, their method is yet to be evaluated with complex tasks such as target driven visual navigation. In this section we point out the challenges when adapting Ma et al.'s model to complex tasks.

The state representation $\phi$ vector mentioned in the USF architecture plays a crucial role. It decouples the scalar reward $r_t$ and learns the USFA $\psi_g^\pi(s_t)$ with the TD-error loss function (Equation 4). The authors Ma et al. [2018b] propose learning $\phi$ using an autoencoder prior to the main reinforcement learning algorithm. $\phi$ is supposed to capture the salient information about each state $s_t$, but when the states consist of complex visual information such as photo-realistic images, defining an optimal $\phi$ representation with an autoencoder can be problematic. Training a convolutional autoencoder which generalize over many states is often prone to over-fitting.

Training $\omega_g$ by a regression loss with respect to a scalar reward $r_t$ (*Equation* 2) can also be problematic. The main reason is that this loss is not informative enough to train $\omega_g$ because during the initial stages of training, the agent will observe very small negative rewards and rarely see the large positive reward goal locations.

$\omega_g$ when decoupled from the scalar reward, captures information about the goal. Ma et al. [2018b] propose to train $\omega_g$ with a separate neural network that uses goal features as input. Training a separate network in our scenario easily leads to over-fitting on the limited number of trained goal locations seen by the agent during training and leads to poor generalization of the agent to new unknown goals.

# 3 Adapting USFA with A3C

Our first contribution is the application of USFA for the complex task of target driven visual navigation. This section introduces how we created a stable architecture while addressing the aforementioned issues.

## 3.1 State Representation

Rather than using a separate network such as an autoencoder to generate $\phi$, we argue it is more beneficial if the agent learns task dependant state representation features while exploring the state space. Thereby, the agent should learn to capture only the salient features relevant to the task of navigation and ignore features that may be solely important for reconstruction. Since in target driven visual navigation, the goal space is a subset of the state space, we used a siamese network to generation both $\phi$ and $\omega_g$.

## 3.2 Reward Prediction Vector with A3C Critic

A major problem with training a stable USFA based DRL agent is the difficulty of training an $\omega_g$ that works successfully with the scalar reward regression loss (*see Equation* 2). In our case, the reward structure is ad-hoc: for every time-step the agent either receives a small negative penalty or a large reward for reaching the goal location. The positive rewards are experienced by the agent much less frequently, particularly at the beginning of training. When training large scale tasks, this class imbalance can be even more detrimental because the reward structure needs to be learnt before the reinforcement learning agent is able to learn a meaningful policy for navigation. If not, this creates an unstable $\omega_g$ which can cause the network to diverge.

To overcome this problem, we propose to exploit the A3C agent's critic update (Value function update). In a conventional A3C algorithm, each agent needs to optimise the critic and policy functions after each episode with the N-step Return [Mnih et al., 2016]. Since the value function can be interpreted by the USFA concept as being a linear combination of $\psi_g(s_t)$ and $\omega_g$, the critic's loss function in an A3C agent can be used to learn $\omega_g$. Unlike training the network with a scalar reward regression loss, this method is more informative because the loss function depends on the episode's discounted scalar rewards. The discounted return calculation for a single episodic step in A3C is depicted in Algorithm 1 in the Supplementary Materials. *Equation* 5 shows how the value function can be decoupled with $\psi$ and $\omega$.

$$\begin{aligned} Loss_{V_{TD}} &= \|r(s_t) + \gamma_t V(s_{t+1}, g) - V(s_t, g)\|^2 \\ &= \|r(s_t) + \gamma_t \psi^\pi(s_{t+1}, g)^\top \omega(g) - \psi^\pi(s_t, g)^\top \omega(g)\|^2 \end{aligned} \quad (5)$$

*Equation* 6 shows the conventional one step TD loss for the SFA branch. It needs to be highlighted that $\psi$ gets updated with both $Loss_{\psi_{TD}}$ and $Loss_{V_{TD}}$.

$$Loss_{\psi_{TD}} = \|\phi(s_t) + \gamma_t \psi_g(s_{t+1}) - \psi_g(s_t)\|^2 \quad (6)$$

To counter the problem of the having only a few training goals to train $\omega$, we utilised the embeddings generated from the Siamese layer as goal information and trained $\omega$ as another branch of the USFA-A3C agent (*see Figure* 5 *in the Supplimentary Materials*).

# 4 Introducing SFDP for USFA

Our second contribution is the addition of a Successor Feature Dependant Policy (SFDP) to the USFA implementation. As mentioned before, $\psi_g(s_t)$ can be seen as an abstract representation of the cumulative sum of the future states the agent will visit by following an optimal policy [Dayan, 1993, Barreto et al., 2017]. Traditionally, successor features are not directly consulted when determining an action [Ma et al., 2018b].

However, we hypothesise that feeding the abstract map of future states could be useful in determining the next action. USF can be described as representing the cumulative sum of discounted future states the agent visits following an optimal policy. This property by itself helps with transfer learning because eventhough different goals have different optimal paths, they can share some common sub-paths. For example, when tasked with finding the microwave and sink in a kitchen, the initial steps of the agent in going to the kitchen will be similar for both tasks. We hypothesised that if the policy has direct access to the USF (*see Equation* 7), the agent will be able to learn from these similar paths. By directly concatenating $\psi_g$ with the final layer of the policy head naively results in $\psi_g$ being updated with gradients from the conventional bellman optimality Equation 3 and the policy gradients of the A3C agent. This can harm the true USF representation and can reduce the transfer learning capabilities of the agent. Therefore in the final model, we stopped the gradient flow from the policy head to the USF branch.

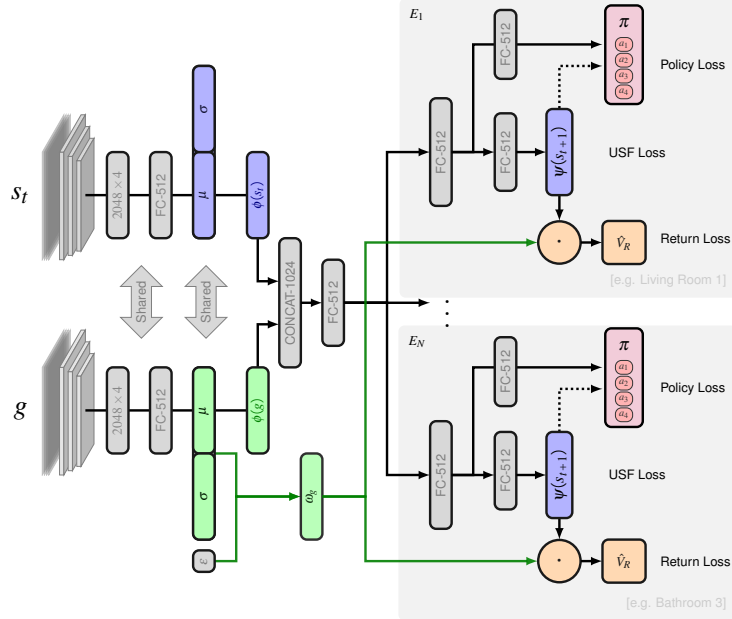

Figure 1: Proposed Network Architecture "VUSFA": The model's input is the current state of the agent $s_t$ and the goal location $g$ as images. These go through a shared simaese encoder $E(z|s_t)$. The reparametrized output $z$ is used to train the $\omega$ vector. The policy is conditioned on the USF vector (dotted line indicates gradients do not flow from policy to the USFA head). The USFA $\psi$ is trained with the temporal difference error using $\phi$ to give the expected future state occupancies. The discounted episode return is used to train both $\omega$ and USFA vectors.

$$\pi(a|s,g;\theta) \leftarrow \pi(a|\psi_t, s_t, g;\theta) \quad (7)$$

The stopping of policy gradients for the USF branch is illustrated in Figure 1 with dotted lines.

## 5 VUSFA

The next modification we made was the introduction of the Variational Siamese Bottleneck (VSB) to improve the quality of USFA and the reward prediction vector. We observed that the embeddings generated by the siamese layers play a key role in improving the overall performance due to their effect on generating $\psi$ and $\omega_g$. We wanted to improve these embeddings without harming stable convergence. Our main hypothesis in selecting the variational information bottleneck was that it will be able to guide the Siamese layers to extract the most informative and meaningful features which will then lead to better generalisation. Having a siamese layer that generates embeddings which produce a robust $\phi$ and $\omega_g$ is key to improving the transfer learning ability of the overall model. If these embeddings are not informative enough, the model can overfit to the training set of goals. To improve the embeddings without harming the training stability, we adapt the concept of Deep Variational Information Bottleneck [Alemi et al., 2016]. Our results show that this addition improves the performance of the overall network.

In the next sections we will describe the theory behind the Variational Information Bottleneck and the training procedure we used in our adaptation of it to the VUSFA agent.

## 5.1 Information Bottleneck

The theory of the Information Bottleneck, introduced by Tishby and Zaslavsky [2015], shows that a deep neural network can be thought of as a trade-off between having a compressed latent representation $Z$ with respect to inputs $X$ while still preserving relevant information with respect to the outputs $Y$. Mathematically, this idea of generating an optimal $Z$ can be achieved using Mutual Information Theorem as follows:

$$\underset{Z}{\text{minimize}}\,[I(X;Z) - \beta I(Y;Z)] \tag{8}$$

Where $I(X;Z)$ is the mutual information between the input features $X$ and the latent representation $Z$ from the hidden layer and $I(Y;Z)$ is the mutual information between output $Y$ and $Z$. Intuitively, the neural network should predict the output while reducing the mutual information between input and the encoding. The minimisation of $I(X;Z)$ encourages the agent to compress the most relevant information about $X$ into $Z$ for the prediction of $Y$.

## 5.2 Deep Variational Information Bottleneck

Deep Variational Information Bottlenecks introduced by Alemi et al. [2016] is a parameterized approach to the Information Bottleneck theory that can be easily used with deep neural networks. This was done by introducing a regularized objective function as follows.

$$J(q(Y|Z), E(Z|X))_{\min} = E_{(Z \sim E(Z|X))}[J(q(Y|Z))] \quad \text{s.t} \quad I(Z,X;\theta) \le I_c \tag{9}$$

Minimizing the loss function $J(q(Y|Z), E(Z|X))$ encourages the neural network to generate an informative compressed embedding $Z$ from the input $X$. *Equation 9* consists of a parametric encoder function $E(Z|X)$ that maps input features $X$ into latent vector $Z$, a decoder function $q(Y|Z)$ that maps $Z$ to output labels $Y$, and a mutual information constraint $I(X,Z) \le I_c$. The generation of $Z$ by the encoder under the Information Constraint $I_c$ can be though of as a bottleneck layer in the network. This bottleneck $Z$ could be applied as any intermediate layer of the neural network.

In our scenario, we applied the Information Bottleneck on the output of the siamese layers (see Figure 1) due to it's direct effects on the estimations of $\pi$, $\psi$, and $\omega$. The siamese layers can therefore be thought of as the encoder $E(Z|X)$. We call this the Variational Siamese Bottleneck (VSB). The VSB enforces an upper-bound $I_c$ on the mutual information term $I(Z,X)$ to encourage the encoder $E(Z|X)$ to focus on the most discriminative features of the input. The encoder needs to generate a latent distribution $Z$ where the mutual information between $X$ and $Z$ does not exceed the scalar Information Constraint $I_c$ where $I_c$ is a hyperparamter.

Since the $I(Z,X;\theta) \le I_c$ term in *Equation 9* is intractable, we cannot directly apply it to a neural network trained with back-propagation. Alemi et al. [2016] introduced a modified version by applying a variational lower bound and a lagrangian multiplier $\beta$ that needs to be updated adaptively. This results in the final deep variational information bottleneck loss in *Equation 10*.

$$J(q,E(Z|X))_{\min} = E_{(z \sim E(Z|X))}[J_p(q(Y|Z))] + \beta E_{(x \sim p(x))}[KL[E(z|x) \| r(z)] - I_c] \tag{10}$$

Since the KL divergence is calculated between two distributions, the encoder outputs the mean and variance of $Z$ from which a sample is taken using the reparametrization trick Kingma and Welling [2013].

We update the Lagrangian multiplier $\beta$ in a similar way to Peng et al. [2018]. $\beta$ gets updated for each actor thread adaptively following *Equation 11*.

$$\beta \leftarrow max(0, \beta + \alpha_\beta(\mathbb{E}_{\mathbf{g} \sim \tilde{p}(\mathbf{g})}[KL[E(\mathbf{z}|\mathbf{g}) \| r(\mathbf{g})]] - I_c) \tag{11}$$

The final loss function of our agent with the Variational Information Bottleneck is shown in the *Equation 12*.

$$J(q,E)_{\min} = E_{(z \sim E(z|x))}[L_{total}] + \beta E_{(x \sim p(x))}[KL[E(z|x) \| r(z)] - I_c] \tag{12}$$

–where $L_{total}$ is the combined loss function $L_{total} = \lambda_\pi \llcorner_\pi + \lambda_\psi \llcorner_\psi + \lambda_V \llcorner_V$. Therefore, the agent needs to minimize both $L_{total}$ and the KL divergence term $[KL[E(z|x) \| r(z)] - I_c]$ at the same time. $\lambda_\pi, \lambda_\psi$ and $\lambda_V$ are hyperparameters.

| Environment | # Trained States | Total States | % States Trained | Model 01 | Model 02 | Model 03 | Model 04 |
|---|---|---|---|---|---|---|---|
| bathroom_02 | 5 | 180 | 2.78% | 14.22% | 20.89% | 22.44% | 27.89% |
| bedroom_04 | 5 | 408 | 1.23% | 17.84% | 20.51% | 20.93% | 23.01% |
| kitchen_02 | 5 | 676 | 0.74% | 10.92% | 11.92% | 11.97% | 17.13% |
| living_room_08 | 5 | 468 | 1.07% | 17.20% | 16.67% | 19.59% | 18.53% |
| All | 20 | 1732 | 1.15% | 15.04% | 16.16% | 17.23% | 20.01% |

Table 1: Zero-shot learning Results: Success rate of the agent reaching all goals within 500 steps without retraining. The agent navigated to each goal location starting from 10 random locations within the simulator. A detailed description of each model can be found in Section 7.

## 6 Network Architecture

Our network (*see Figure 1*) takes the four most recent states the agent has visited as $s_t$ and four repeated goal states as the $g$. Then the resnet-50 embeddings related to both $s_t$ and $g$ go through a siamese encoder and generates the mean and variance vectors, which are the parameters of the latent distribution Z. The $\phi$ embeddings with respect to the goal $g$ then go through a fully connected layer to predict the goal coefficients $\omega$ vector. The network's last part predicts the policy $\pi(a_t|s_t, g, \psi)$ and the USFA $\psi_g(s_t)$. Similar to Zhu et al. [2017], we include a separate policy ($\pi$), USFA $\psi$ and the expected sum of future rewards prediction heads for each scene (e.g. Bathroom) Kolve et al. [2017].

### 6.1 Training VUSFA

The training procedure for our model is based on the A3C algorithm and is shown in *Algorithm 1*. The reparameterized embedding was not directly used in predicting the policy $\pi$ and the USFA $\psi$ to maintain a stable procedure. Instead, the mean vectors of the state representation $\phi$ for both goal and state were used. These mean vectors were concatenated together and fed through the layers used for predicting the policy and USFA as shown in *Figure 1*. We used the reparameterized embeddings from the bottleneck layer to predict $\omega$ since the $\omega$ vector is the most important element in the USF architecture that decouples the value function. The objective behind this reparametrization was to make create an $\omega$ that is robust and generalizable that would not easily overfit. During inference, we use reparameterized values for both goal and state encoding which we assume added more generalizability and exploration and improved zero-shot navigation of the agent.

## 7 Experimental Evaluation

The evaluation of the agent under the task of target driven visual navigation has been conducted in two ways. First, the agent was evaluated on its zero-shot learning ability. The second evaluation criteria was the time taken for the agent to adapt to new unknown goals when fine-tuning. Both evaluation criteria belong to the domain of Transfer in Reinforcement Learning [Taylor and Stone, 2009] and will be described in the following sections. Prior to evaluation, all models were trained on four scenes for 20 different goals until convergence. We took the deep Siamese A3C model by Zhu et al. [2017] as the baseline, since it is the most relevant work done using the AI2THOR simulator Kolve et al. [2017].

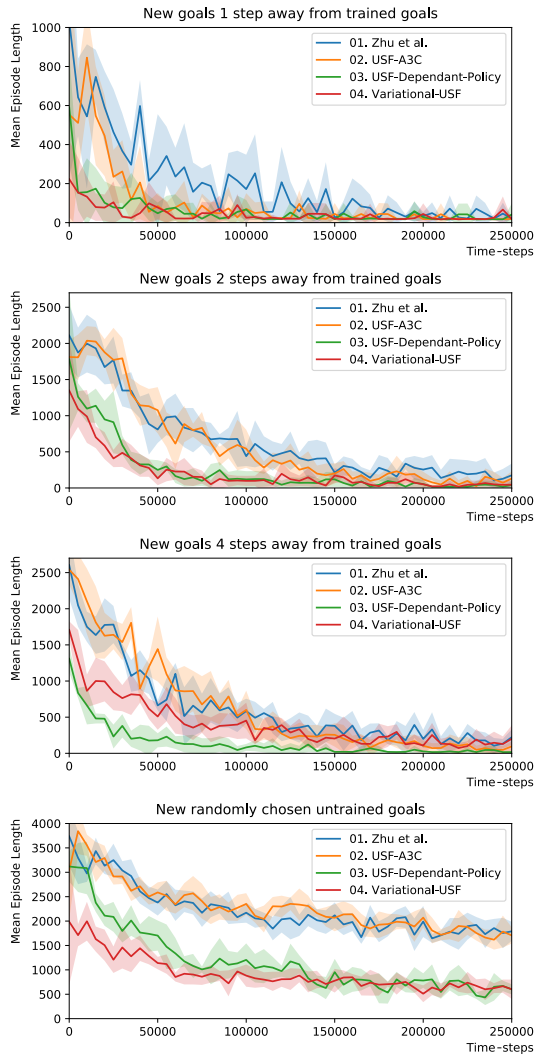

Figure 2: Agent's transfer learning ability: No. of training time-steps plotted against the average length of an episode. Shorter episode lengths indicate the agent has learnt to navigate to goals in a lower no. of steps (shaded area is the standard deviation over 100 time-steps).

Moreover, we were also successful in training the agent from scratch with a CNN (replacing the resnet features). Although adding an LSTM results in further performance increases, we used resnet features instead. We decided to do so to keep our training time low and the evaluation consistent.

We evaluated all variations of our proposed model in a benchmark with the state-of-the-art:

**Model 01** : Implementation of Zhu et al. [2017]'s model using GVF

**Model 02** : Using USFA

**Model 03** : Adding SFDP to Model 02 (see Figure 5 in Supplementary Materials)

**Model 04** : Adding VSB to Model 03 (we call this *VUSFA*)

### 7.1 Zero-Shot Navigation Ability

The aim of zero-shot navigation is to see weather the agent is be able to reach a wide range of goals while being trained on only a very limited subset of goals. In particular, the zero-shot learning capability of the agent was evaluated by testing the agent's average successful attempts to find new goal locations. In the evaluation process, we follow a similar criteria to Zhu et al. [2017], in which we tested whether the agent was able to reach the goal in less than 500 time-steps. We constituted this as a successful episode. We evaluated the success rate of reaching all goals in each environment. We repeated this procedure 10 times (trials), in which the agent always started from a random location. We trained our models on only 20 goal states spread evenly across four environments using the AI2THOR simulator. This represents less than 1.2% of the total number of states. In-spite of this, even the worst performing model was able to generalize to over 16% of all states.

Table 1 shows that all proposed algorithms (Model 02–04) are able to successfully reach more locations without training than the baseline Model 01. The USFA-based policies consistently generalise better than Zhu et al. [2017].

### 7.2 Transfer Learning

It can take a large amount of time (in the order of several days), to train a complex DRL agent. Therefore, it is impractical to re-train agents when the task slightly changes. Instead, the agent should be able to use previous knowledge to adapt to new tasks quickly.

We evaluated the transfer learning ability of all four models to 20 new goals. In order to evaluate how the closeness of the new goals effect the agent's performance, we tested the models on states that are 1, 2, and 4 steps away from already trained goals as well as completely random goals. We sampled 5 random states from each environment, excluding the already trained goals to get the new goals. We used 5 trials, meaning repeating this process 5 times with random states which are different to the previously learned ones. To ensure a fair comparison, we kept the random seeds constant between the models.

*Figure* 2 shows the number of time-steps required for the model to adapt to new goals. It becomes clear that the USFA-based policies are consistently able to decrease the number of steps taken to reach the goal faster than the baseline model. Moreover, using the SFDP with USFA resulted in a further decrease in time-steps required and thus showed to have a positive effect on the model's transfer learning ability. As shown in *Figure* 2, VUSFA is usually able to further improve performance.

## 8 Conclusion & Future Work

We proposed Variational Universal Successor Features Approximator (*VUSFA*) to solve rather complex tasks, such as target driven visual navigation in photorealistic environments using the AI2THOR simulator. To our knowledge, this is the first time the Deep Variational Information Bottleneck theory has been applied with Universal Successor Features in Deep Reinforcement Learning. Our results indicate that *VUSFA* is able to improve the transfer learning ability in respect to previous state-of-the-art GVF and USF-RL based research . Our approach is generalizable and can be easily adapted to various tasks other than navigation. For re-implementation, we provide the source code via our github repository[1]. Our approach introduces a new perspective and should be considered in future research aiming to improve transfer learning for Deep Reinforcement Learning. In particular, further research could look into exploration of the semantical impacts of $\phi$, $\omega$, and $\psi$.

# Supplementary Materials

## A  Algorithm

**Algorithm 1** Variational A3C-USF Algorithm pseudocode for each actor thread
---
Assuming global parameter vectors for value function and policy as $\theta_\pi$, $\theta_\psi$ and $\theta_\omega$
Assuming global shared counter as T = 0
Assuming thread specific parameter vectors for value function and policy as $\theta'_\pi$, $\theta'_\psi$ and $\theta'_\omega$ d
Initialize thread step counter $t \leftarrow 1$
1: **repeat**
2:    Reset gradients : $d\theta_\pi \leftarrow 0, d\theta_\psi \leftarrow 0, d\theta_\omega \leftarrow 0$
3:    Synchronise thread specific parameters with global network $\theta'_\pi = \theta_\pi, \theta'_\psi = \theta_\psi, \theta'_\omega = \theta_\omega$
4:    $t_{start} = t$
5:    Get initial state $s_t$, goal $g$
6:    **repeat**
7:       Perform an action $a_t$ according to the current policy $\pi(a_t|s_t, g : \theta'_\pi)$
8:       Receive the scalar reward $r_t$, the new state $s_{t+1}$
9:       Collect roll-outs $[s_t, r_t, s_{t+1}]$
10:      $t \leftarrow t + 1$
11:      $T \leftarrow T + 1$
12:   **until** terminal $s_t$ or $t - t_{start} == t_{max}$
13:   Bootstrapping the return R from last state of the episode
14:   $R_\psi = \begin{cases} 0 & s_t = terminal \\ \psi(s_t, g; \theta_\psi) & s_t \neq terminal \end{cases}$
15:   $R_V = \begin{cases} 0 & s_t = terminal \\ \psi(s_t, g; \theta_\psi)\omega(g; \theta_\omega) & s_t \neq terminal \end{cases}$
16:   **for** $i \in t-1, ..., t_{start}$ **do**
17:      $R_\psi \leftarrow \psi(i, g; \theta_\psi) + \gamma R_\psi$
18:      $R_V \leftarrow \psi(s_t, g; \theta_\psi)\omega(g; \theta_\omega) + \gamma R_V$
19:      $A_V = R_V - Vi$
20:      Collect $[R_v, R_\psi, A_V]$
21:   Compute $\beta \leftarrow max(0, \beta + \alpha_\beta(\mathbb{E}_{\mathbf{g} \sim \tilde{p}(\mathbf{g})}[KL[E(\mathbf{z}|\mathbf{g})\|r(\mathbf{g})]] - I_c)$ where $I_c = 0.2$ is a hyperparameter
22:   Compute $\llcorner_{KL} = \phi_\mu(g)^2 + \phi_\sigma(g)^2 - log(\phi_\sigma(g)^2) - 1$
23:   Compute $\llcorner_\psi = \|R_\psi - \psi(s_t, g; \theta_\psi)\|^2$
24:   Compute $\llcorner_V = \|R_V - \psi(s_{t+1}, g; \theta_\psi)\omega(g; \theta_\omega)\|^2$
25:   Compute $\llcorner_\pi = \log[(\pi(s_t, g; \theta_\pi)]A_V$
26:   Calculate gradients on loss $\lambda_\pi \llcorner_\pi + \lambda_\psi \llcorner_\psi + \lambda_V \llcorner_V + \beta \llcorner_{KL}$
27:   Computing $d\theta_\psi, d\theta_\pi$ and $d\theta_\omega$, perform asynchronous update of $\theta_\psi, \theta_\pi$ and $\theta_\omega$
28: **until** $T > T_{max}$

## B  Implementation Details

We used the fundamental A3C framework to train our agent. The initial A3C agent contains 100 threads assigned to 20 goals (5 goals from each scene). The 20 goals used for training can be seen in Figure 3. For the transfer learning tasks, 20 new goals were randomly selected excluding the trained goals. An ideal trajectory of an agent is illustrated in the Figure 4.

For training, we used a high performance computer with 4x Xeon 6136 @3Ghz Processor (total of 48 cores); Memory - 1.18TB, GPU - Tesla V100 with 32GB memory per GPU.

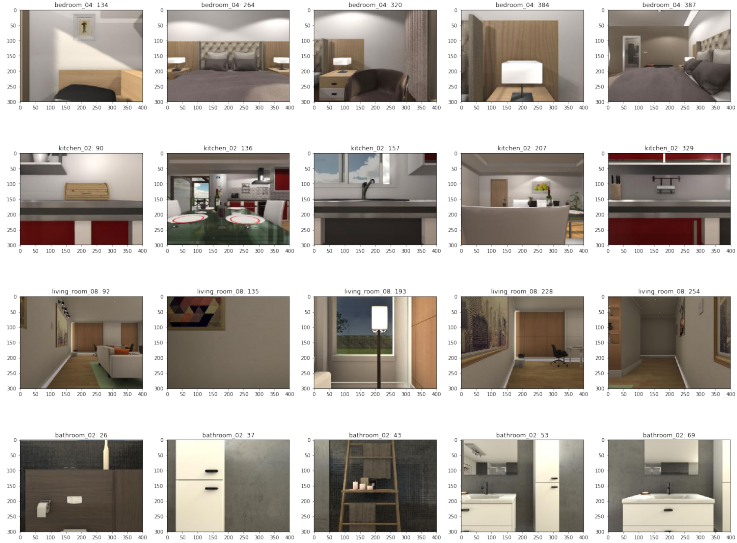

Figure 3: The twenty goal locations used for training

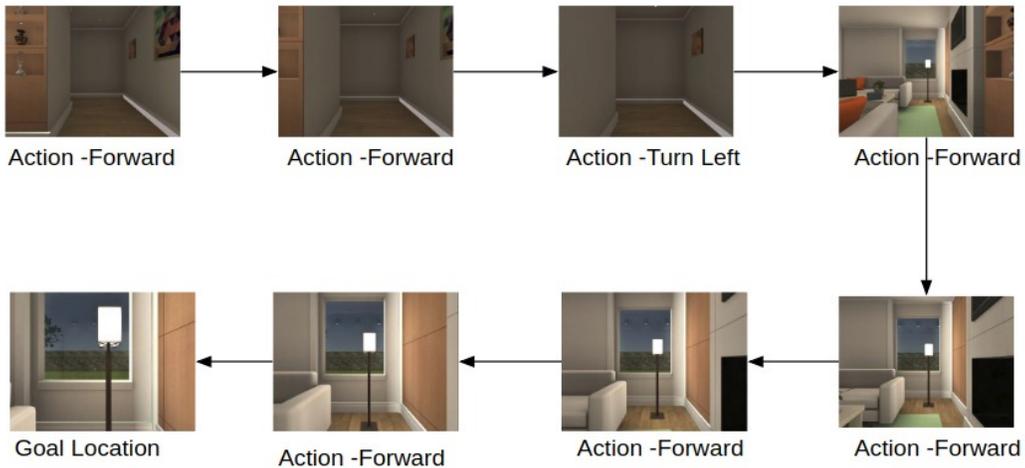

Figure 4: Example of a trajectory of an agent inside the AI2THOR simulator

### B.1 Hyperparameter Tuning

For the proposed method, we varied the weight of the USF TD error loss, KL loss and the information constraint $I_c$.

We did limited hyperparameter tuning as detailed in Table 2. $\lambda_V = 0.5$, $\lambda_\psi = 0.0005$ and $I_c = 0.2$ gave good results and were used in the plots.

| Parameters | Values Explored | Values Used |
|---|---|---|
| $I_c$ | 0.2,0.5,0.1 | 0.2 |
| $\lambda_\psi$ | 0.0005,0.0006,0.0009 | 0.0005 |
| $\lambda_V$ | 0.5 | 0.5 |
| $\lambda_\pi$ | 1.0 | 1.0 |

Table 2: Explored Hyperparameters

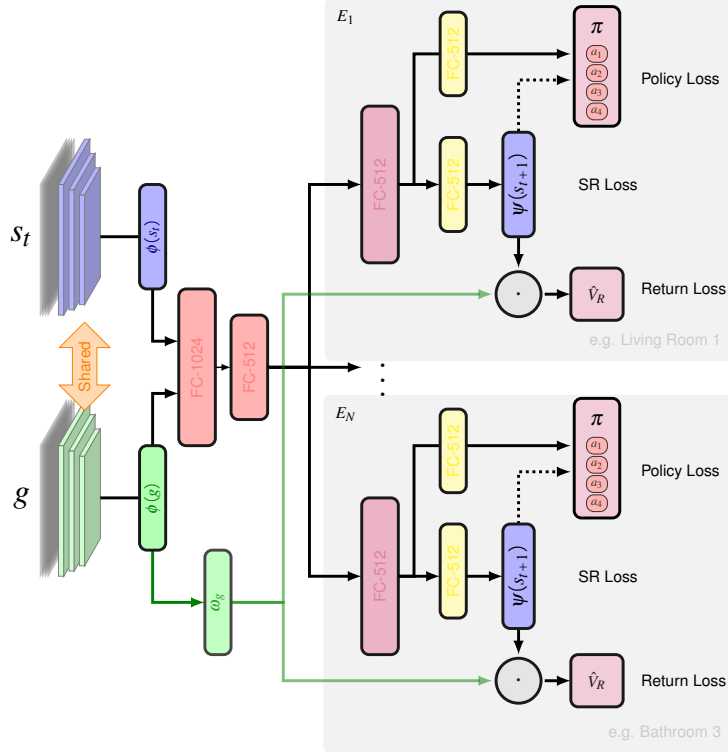

Figure 5: Architecture used for the SFDP-A3C model
The input to the model is the current state of the agent $s_t$ and the goal location $g$ as images. These go through a shared siamese layer to generate embeddings for the goal and the state. The policy is conditioned on the USF vector (dotted line indicates gradients do not flow from policy to the USFA head). The USFA $\psi$ is trained with the temporal difference error using $\phi$ to give the expected future state occupancies. The discounted episode return is used to train both $\omega$ and USFA vector.

## C  Related Work on GVF and USF

Designing a value function that is capable of adapting to different tasks is based on a concept known as General Value Functions (GVF) [Sutton et al., 2011]. An extension to this idea, called Universal Value Function Approximators (UVFAs), was introduced by Schaul et al. [2015]. The core idea of UVFAs is to represent a large set of optimal value functions by a single, unified function approximator that generalises over both states and goals. Although theoretically sound, learning a sufficiently working UVFA is a challenging task [Ma et al., 2018b].

Successor Representation (SR) [Dayan, 1993] emerged from the field of cognitive science and is modelled on the brain's capability to create a reusable predictive map. Recently, SR was combined with Deep Learning to create Deep Successor Reinforcement Learning by [Kulkarni et al., 2016]. Based on Deep Q-Network (DQN) [Mnih et al., 2013] fundamentals, Kulkarni et al. [2016] were able to learn task-specific features that were able to quickly adapt to distal changes in the reward by fine-tuning only the reward prediction feature vector. Transfer in RL has been evaluated on multiple similar tasks by [Barreto et al., 2017], who introduced Successor Features (SF). They adapted SR to be applicable in the continuous domain and were able to show how classic policy improvement can be extended to multiple policies. Extending Barreto et al. [2017] to Deep Learning, Barreto et al. [2018] additionally showed how these models can be stably trained. For the problem of visual navigation, a SR-based DRL architecture similar to Kulkarni et al. [2016] was used by Zhang et al. [2017]. Unlike our approach, they showcase their solution in simple maze-like environments using DQN as the baseline method, while we use actor-critic methods in a photorealistic simulation environment. DQN-based techniques frequently suffer from stability issues when applied to complex problems, such as large-scale navigation [Barreto et al., 2018].

USFA [Ma et al., 2018b] is the most recent extension to SR. Unlike previous methods, which were based on DQN, USFA learns a policy directly by modeling it with actor-critic methods. Similar to SR, USFA modifies the policy with successor features. DQN-based approaches learn an optimal action-value function indirectly. USFA attempts to obtain a GVF which can be directly used to obtain an optimal policy. It can be seen as a combination of the SF and UVFA methods as discussed earlier. Unlike methods based on DQN, USFA is able to directly optimize an agent to learn multiple tasks simultaneously. However, USFA has not been tested on high-dimensional complex problems. This paper shows the adaption of USFA to more complex task for target driven visual navigation in a photorealistic environment.

Borsa et al. [2018] also extended the idea of SF-RL in to USFA by adapting the concepts of GVF and generalized policy improvement theorem. [Borsa et al., 2018] work describes a USFA training method by adapting $\varepsilon$-greedy Q learning for set of simple tasks in Deepmind-Lab [Beattie et al., 2016] simulator. We found their method is hard to adapt to our problem domain mainly due to defining the state representation vector as a linear combination of a set of tasks. This method can work with a simple set of known tasks such as collecting different objects in the simulator. But in our case, each new task is similar to a new goal location selected from the state space. Also, the authors have not extended this algorithm to work with policy gradient methods.

## D  Extended Abstract on Universal Successor Feature Approximators

Similar to the UVFA [Schaul et al., 2015], a goal-dependant SF is approximated with Universal Successor Feature Approximators (USFA) [Ma et al., 2018b, Borsa et al., 2018]. The goal dependant USF obeys the bellman update rule similar to SF (Equation 13).

$$\psi^\pi(s_t, g) = \phi_{t+1} + \gamma E^\pi[\psi^\pi(s_{t+1}, g)|s = s_t)] \quad (13)$$

### D.1  Decoupling the Goal Oriented Value function with Universal Successor Features

Similar to Successor Feature based DRL methods [Kulkarni et al., 2016], the decoupled value function with USF can be written as Equation 14.

$$\begin{aligned} V_g^\pi(s) &= \mathbb{E}^\pi \left[ \sum_{t=0}^\infty \phi(s_t, a_t, s_{t+1}; \theta_\phi) \prod_{k=0}^t \gamma_g(s_k) \middle| s_0 = s \right]^\top \omega(g_t; \theta_\omega) \\ &= \psi^\pi(s_t, g_t; \theta_\psi)^\top \omega(g_t; \theta_\omega) \end{aligned} \quad (14)$$

In Equation 14 $\psi^\pi(s_t, g_t; \theta_\psi)$, $\omega(g_t; \theta_\omega)$ and $\phi(s_t, a_t, s_{t+1}; \theta_\phi)$ should generalize over both goals and states.

The fundamental architecture to learn USF was proposed by Ma et al. [2018b] and is shown in Figure 6. The architecture consists of three sub-networks that need to be trained. The first one is the state representation network with parameters $\Phi$. This network is an autoencoder which gets trained before the training of the DRL agent. Then there are two heads to approximate the USF ($\theta_\pi$), and policy ($\theta_\psi$). These two heads share parameters in the early layers with a fork in the last fully connected layer. Finally, $\theta_w$ is learned. This final reward prediction network should be able to generate $\omega$ for different goals which will later be used to train the USF agent.

It is important to compare the architectural differences between the USF-RL methods [Ma et al., 2018b] with the previous DQN based SF-RL methods [Kulkarni et al., 2016, Zhang et al., 2017]. Unlike in SF-RL, $\phi$ is not learned simultaneously while training the DRL agent. The $\phi$ prediction network in the USF-RL agent is trained with an autoencoder separately prior to the training phase. In contrast to the SF-RL methods which stores an $\omega$ vector for each task, in USF-RL the $\omega$ vector is predicted with a network that takes information about the goal as the input. Maintaining a separate network to generate $\omega$ given goal information gives the luxury for the agent to quickly adapt to novel goals.

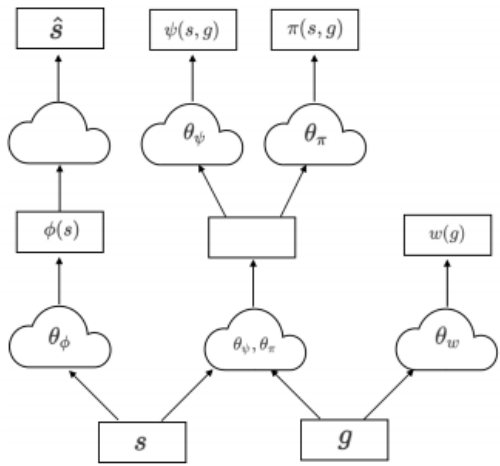

Figure 6: Proposed architecture by Ma et al. [2018b] to train an actor-critic agent with USF (image from Ma et al. [2018b])

# E Mathematical framework for training the Varational Information Bottleneck with Dual Gradient Descent

The theory of Information Bottleneck (IB) by Tishby and Zaslavsky [2015] explains a method of designing a neural network considering the trade-off between accuracy and complexity. The theory of the IB can be described with respect to supervised learning as follows:

Assume there are set of classes $Y$, a multi-dimensional feature vector $X$. The network's goal is to compress $X$ to a "less complex" representation $Z$ that shares enough information to predict $Y$ correctly. The concept of IB describes the fluctuation between compression versus preserving relevant information in $Z$. Mathematically, the idea of generating an optimal $Z$ can be presented as described in Equation 15.

$$\text{minimize} \quad I(X,Z) - \beta I(Y,Z) \tag{15}$$

In Equation 15 the term $I(X,Z)$ is the mutual information between the features $X$ and the hidden layer $Z$. $I(Y,Z)$ represents the mutual information between output $Y$ and the hidden layer $Z$. Intuitively, Equation 15 says that the neural network should predict the label while reducing the mutual information between the input and the encoding. Reducing of the $I(X,Z)$ term encourages the agent to compress the $X$ into $Z$ where $Z$ consists of only the most important information about the input features. The $\beta$ is the trade-off hyperparameter which controls the quality of $Z$.

The mutual information term $I(X,Z)$ between input features and the encoder embeddings can be illustrated in Equation 16.

$$\begin{aligned} I(X,Z) &= \int p(x,z) \log(\frac{p(x,z)}{p(x)p(z)}) dx dz \\ &= \int p(x) E(z|x) \log(\frac{E(z|x)}{p(z)}) dx dz \end{aligned} \tag{16}$$

In Equation 16, the joint distribution of $p(x,z)$ can be illustrated with the encoder $E(z|x)$ and the input distribution $p(x)$. Another important factor in Equation 16 is the distribution of the latent variable $p(z)$ is intractable. So the $p(z)$ term is replaced with a known prior distribution $r(z)$. This assumption introduces an upper-bound to the mutual information term $I(X,Z)$ as mentioned in Equation 17.

$$I(X,Z) \leq \int p(x) E(z|x) \log(\frac{E(z|x)}{r(z)}) dx dz \\ \leq E_{(x \sim p(x))} [KL[E(z|x) \| r(z)]] \qquad (17)$$

In Equation 17, the sum of the probability distribution of input $X$ can be transformed into an expectation operation. The expectation operation simplifies the $I(X,Z)$ into a KL divergence between the distribution of $Z$ generated by the parametric encoder and the approximated prior $r(z)$. Modification of the $I(X,Z)$ with the KL divergence makes it easier to train a neural network with the loss function mentioned in Equation 9.

The interpretation of $I(X,Z)$ with KL divergence allow us to modify the conditional loss function (Equation 9) as follows in Equation 18.

$$J(q,E)_{\min} = E_{(z \sim E(z|x))} [J_p(q(y|z))] \quad \text{s.t} \quad E_{(x \sim p(x))} [KL[E(z|x) \| r(z)]] \leq I_c \qquad (18)$$

The constraint term of $E_{(x \sim p(x))} [KL[E(z|x) \| r(z)]] \leq I_c$ mentioned in Equation 18 can be subsumed into the loss function $J(q,E)_{\min}$ with a Lagrangian Multiplier [Bertsekas, 2014] $\beta$ that needed to be updated in an adaptive manner with the training procedure of the neural network.

Moreover, Alemi et al. [Alemi et al., 2016] evaluated the method of supervised learning tasks and showed that the models trained with a VIB could be less prone to be overfitting and more robust to adversarial examples. In this paper, we successfully illustrated the adoption of the concept of Deep VIB with the USF to substantially improve the transfer learning ability of a navigation agent.